\documentclass[10pt,twocolumn,letterpaper]{article}

\usepackage{iccv}
\usepackage{times}
\usepackage{epsfig}
\usepackage{graphicx}
\usepackage{amsmath}
\usepackage{amssymb}

\usepackage{booktabs}
\usepackage{balance}
\usepackage[normalem]{ulem}

\usepackage[pagebackref=true,breaklinks=true,letterpaper=true,colorlinks,bookmarks=false]{hyperref}

\usepackage[capitalize]{cleveref}
\crefname{section}{Sec.}{Secs.}
\Crefname{section}{Section}{Sections}
\Crefname{table}{Table}{Tables}
\crefname{table}{Tab.}{Tabs.}

\iccvfinalcopy 

\def\iccvPaperID{5204} 

\ificcvfinal\fi

\begin{document}

\title{iBARLE: imBalance-Aware Room Layout Estimation}

\author{Taotao Jing$^1$, \; Lichen Wang$^2$, \; Naji Khosravan$^2$, \; Zhiqiang Wan$^2$, \; Zachary Bessinger$^2$, 
\\
Zhengming Ding$^1$, \; Sing Bing Kang$^2$ \\
$^1$Tulane University \;\;\;\; $^2$Zillow Group 
\\
{\tt\small \{tjing,zding1\}@tulane.edu}
\\
{\tt\small \{lichenw,najik,zhiqiangw,zacharybe,singbingk\}@zillowgroup.com }
}

\maketitle
\ificcvfinal\thispagestyle{empty}\fi

\begin{abstract}
Room layout estimation predicts layouts from a single panorama. It requires datasets with large-scale and diverse room shapes to well train the models. However, there are significant imbalances in real-world datasets including the dimensions of layout complexity, camera locations, and variation in scene appearance. These issues considerably influence the model training performance. In this work, we propose imBalance-Aware Room Layout Estimation (iBARLE) framework to address these issues. iBARLE consists of: (1) Appearance Variation Generation (AVG) module, which promotes visual appearance domain generalization, (2) Complex Structure Mix-up (CSMix) module, which enhances generalizability w.r.t. room structure, and (3) a gradient-based layout objective function, which allows more effective accounting for occlusions in complex layouts. All modules are jointly trained and help each other to achieve the best performance. Experiments and ablation studies based on ZInD~\cite{cruz2021zillow} dataset illustrate that iBARLE has state-of-the-art performance compared with other layout estimation baselines.

\end{abstract}




\section{Introduction}\label{sec:intro}
With the recent advancements in computer vision related applications (e.g., AR/VR, virtual touring, and navigation), room layout estimation is receiving a lot of attention from researchers. Specifically, panorama-based layout estimation has been a major area of focus due to the increased $360^{\circ}$ field of view~\cite{zhang2014panocontext}. A great deal of progress was made on monocular layout estimation based on a single panorama~\cite{sun2019horizonnet, yang2019dula, sun2021hohonet}. Some directly use the equirectangular panorama \cite{pintore2020atlantanet} while others combine the equirectangular panorama with its perspective top-down view \cite{wang2022psmnet}. A recent trend of papers formulate the problem as a 1-dimension sequence that represents depth on the horizon line of the panorama and calculates the room height by the consistency between the horizon-depth of ceiling and floor boundaries \cite{wang2021led}. Some methods directly predict the room height to make better geometry awareness of the room layout in the vertical direction \cite{jiang2022lgt}.

\begin{figure}[t]
\centering
\includegraphics[width=1\linewidth]{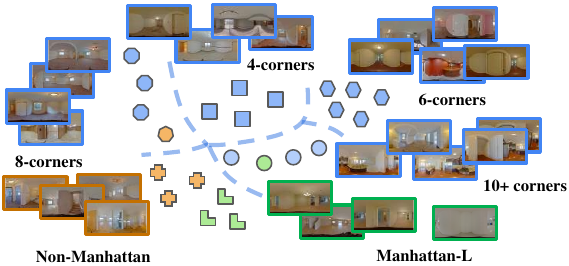}
\caption{Imbalanced and diverse sample distributions such as various room corner numbers, appearances, Manhattan styles, and capture locations which downgrade estimation performances.}
\label{fig:imbalance}
\vspace{-4mm}
\end{figure}

However, these techniques are less effective for complex room shapes (e.g., self-occlusion and non-Manhattan)~\cite{wang2022psmnet}. As a result, the majority of these approaches conform to the Manhattan World or Atlanta World assumption, as well as their corresponding post-processing strategies, resulting in promising performances for simple rooms, while failing to achieve the same level of performance for complex layouts. For instance, ZInD~\cite{cruz2021zillow} is a very large-scale indoor dataset. The performance of most single-panorama layout estimation solutions degrades in complex room shape scenarios. On top of that, there are implicit data imbalance challenges present in real-world datasets such as different camera poses, illumination changes, and texture variations (see Figure~\ref{fig:imbalance}). As shown in Figure~\ref{fig:dataset_split}, simple rooms with only four corners take the majority of the dataset while complex rooms with nine corners make up only $2\%$ of the dataset. Consequently, models trained on such datasets will have a tendency to get biased towards the majority.


In this work, we address the data imbalance and appearance variation issues associated with room layout estimation. We specifically design an Appearance Variation Generalization (AVG) module to overcome appearance variations that are present in real-world datasets (e.g., pose, texture, illumination), a Complex Structure Mix-up (CSMix) module to handle the long tail imbalanced distribution of structural complexity of the training and test data, and a layout gradient-based cost function to better deal with occlusions in complex indoor spaces. The contributions of our work are listed below:
\begin{itemize}
\vspace{-2mm}
\item To the best of our knowledge, our work is the first to tackle the data imbalance issue in layout estimation systematically and based on different sources of bias in the indoor dataset.
\vspace{-3mm}
\item Our framework, iBARLE, introduces architectural and training novelties through our proposed Appearance Variation Generalization (AVG) module, Complex Structure Mix-up (CSMix) module, and our layout gradient-based cost function.
\vspace{-3mm}
\item We show iBARLE consistently achieves the state-of-the-art performance for both simple/conventional and complex real-world indoor layout estimation tasks through experiments and ablation studies.
\end{itemize}

\section{Related Work}\label{sec:related}
\paragraph{Indoor Panoramic Layout Estimation.}

Most panorama-based indoor layout estimation efforts use simple indoor room shapes with Manhattan World~\cite{coughlan1999manhattan} assumptions and Atlanta World \cite{schindler2004atlanta} assumptions and corresponding pose-processing operations. Specifically, convolutional neural networks (CNNs) take panorama images as input to extract visual features which will be used to estimate layout. With LayoutNet, layout estimation of a cuboid room with Manhattan constraints is reconstructed based on images aligned with vanishing points and detected layout elements such as corners and boundaries, and the model designed for cuboid rooms is further extended to predicting Manhattan layouts in general \cite{zou2018layoutnet}. In contrast, Dula-Net projects the panorama images into two different views, equirectangular and perspective, to predict the floor and ceiling probability maps and two-dimensional floor plans \cite{yang2019dula}. 

A new dataset containing panoramas of Manhattan-world room layouts with different numbers of corners is introduced as Realtor360 to learn more complex room layouts \cite{yang2019dula}. Moreover, HorizonNet outperforms other strategies by representing a 2D room layout as three 1-D vectors at each image column, and the 1-D sequence vectors encode the positions of floor-wall and ceiling-wall, and the existence of wall-wall boundaries \cite{sun2019horizonnet}. Although the two improved frameworks LayoutNet v2 and Dula-Net v2 achieve better performance on cuboid datasets, HorizonNet has been demonstrated to be more effective on the new MatterportLayout dataset for general Manhattan layout estimation tasks~\cite{zou2021manhattan}. AtlantaNet breaks through the Manhattan World limitations and projects the original gravity-aligned panorama images on two horizontal planes to reconstruct the Atlanta World 3D bounding surfaces of the rooms~\cite{pintore2020atlantanet}. 

HoHoNet is the first work exploring compact latent horizontal features learning for efficient and accurate layout reconstruction and depth estimation equipped with Efficient Height Compression (EHC) and multi-head self-attention (MSA) modules \cite{sun2021hohonet}. PSMNet is a pioneering end-to-end joint layout-pose deep architecture for large and complex room layout estimation from a pair of panoramas \cite{wang2022psmnet}. LED$^2$-Net goes beyond conventional 3D layout estimation by predicting depth on the horizon line of the panorama, and a differentiating depth rendering procedure is proposed to maximize the 3D geometric information without needing to provide the ground-truth depth \cite{wang2021led}. LGT-Net further extends the LED$^2$-Net with an SWG-Transformer module, which consists of shifted window blocks and global blocks, to predict both horizon depth and room height with a planar geometry aware loss to supervise the estimation of the planeness of walls and turning of corners \cite{jiang2022lgt}.

\paragraph{Imbalanced Data Training.}
Due to the widespread application of artificial intelligence systems in our daily lives, the issue of data imbalance has gained increasing attention in recent years~\cite{mehrabi2021survey}. Research about imbalance AI seeks to ensure that AI systems make decisions and predictions without discriminating against certain groups of data when making crucial and life-changing decisions. The algorithms that target bias generally fall into three categories: (1) Pre-processing methods attempt to remove the underlying discrimination from the data \cite{bellamy2019ai, d2017conscientious}. (2) In-processing techniques seek to modify the learning and training strategies, either by incorporating changes into the objective function or imposing a constraint, to eliminate discrimination during model training \cite{bellamy2019ai, berk2017convex}. (3) Post-processing is performed after training stage if the algorithm treats the AI model as a black box without modifying the training data or learning strategies. Specifically, a hold-out set that was not directly involved in the training process is used during the post-processing phase to reassign the initial labels assigned by the black-box model \cite{bellamy2019ai, berk2017convex}.

\begin{figure*}[t]
\centering
\includegraphics[width=0.98\linewidth]{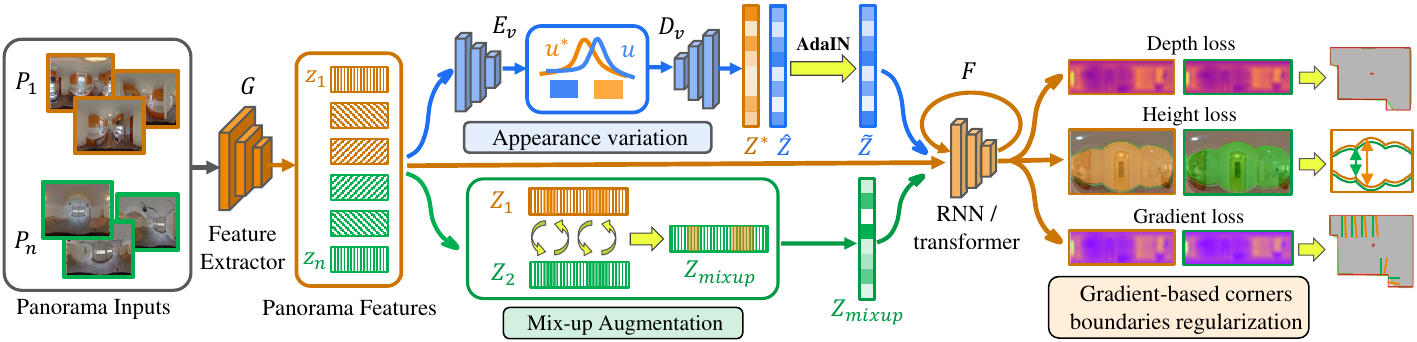}
\vspace{-1mm}
\caption{Framework of iBARLE. There are four core modules: visual feature extractor $G(\cdot)$, transformer-based sequential layout estimation $F(\cdot)$, Appearance Variation Generalization (AVG), and Complex Structure Mix-up (CSMix). AVG enhances the generalization capability by enhancing invariance to appearance changes. CSMix improves the prediction of complex room shapes by generating more diverse layouts that improves the balance of the overall sample distributions.}
\label{fig:framework}
\vspace{-3mm}
\end{figure*}

\paragraph{Domain Generalization.}
As part of the quest to develop models that can generalize to unknown distributions, Domain Generalization (DG), which refers to out-of-distribution generalization, has attracted increasing attention in recent years \cite{wang2022generalizing}. A majority of existing domain generalization methods can be classified into three types: (1) Data manipulation focuses on manipulating the inputs in order to assist in the learning of general representations. Data augmentation \cite{shorten2019survey, honarvar2020domain} and data generalization \cite{qiao2020learning, li2021progressive, rahman2019multi} are two types of popular techniques in this regard. Additionally, mix-up on the original image-level \cite{wang2020domainmix, wang2020heterogeneous} or feature-level \cite{zhou2020domain, xu2021fourier} is a popular and effective technique. (2) Representation learning \cite{bengio2013representation} consists of two representative techniques: a). Learning domain-invariant representations via kernel functions \cite{grubinger2015domain, ben2006analysis}, adversarial learning \cite{ganin2015unsupervised, ganin2016domain}, feature alignment \cite{motiian2017unified, jin2020style, li2018domain}, etc.; b). Feature disentanglement is the process of distancing features into domain-shared and/or domain-specific elements to enhance generalization \cite{khosla2012undoing, niu2015multi, ding2017deep}. (3) Learning strategies that promote generalization employ general strategies such as ensemble learning \cite{d2018domain, wu2021collaborative, mancini2018best}, meta-learning \cite{finn2017model, snell2017prototypical, santoro2016meta}, gradient operations \cite{shi2022gradient, huang2020self}, and self-supervised learning \cite{kim2021selfreg, jeon2021feature, li2021domain}. To the best of our knowledge, we are the first to apply the domain generalization techniques to address the appearance variation issues associated with room layout estimation.



\section{Framework Overview}\label{sec:overview}
The iBARLE framework, illustrated in Figure~\ref{fig:framework}, has four core modules: visual feature extractor $G(\cdot)$, transformer-based sequential layout estimation module $F(\cdot)$, Appearance Variation Generalization module (AVG), and Complex Structure Mix-up module (CSMix). The room layout is estimated by predicting the horizon depths and heights of multiple points sampled from the floor boundary of the polygon of the panorama image \cite{wang2021led, jiang2022lgt}. During the training phase, the AVG and CSMix modules are used to compensate for the imbalance in the data; during the test phase, only the feature extraction and the sequential layout estimation module are called to predict the room layout.

\section{Proposed Algorithm}\label{sec:algorithm}
In this section, we first go over the overall framework to predict the room layout based on a single panorama input. We then introduce the two modules, Appearance Variation Generalization (\textbf{AVG}) and Complex Structure Mix-up (\textbf{CSMix}). Finally, the gradient-based \textit{Corners and Occlusions objective} is presented. It improves the efficiency of the model in detecting visible wall-wall corners and occlusion boundaries in complex spatial arrangements.


\subsection{Framework}
Reconstructing the 3D indoor layout by predicting the horizon depth and room height is an effective strategy that is widely adopted by recent methods~\cite{jiang2022lgt, wang2021led}. Specifically, N points $\mathbf{P} = \{\mathbf{p}_i\}_{i=1}^{N}$ with equal longitude interval are sampled from the floor-boundary of the polygon of the panorama image, where the longitudes of the sampled points are $\{ \theta_i = 2\pi (\frac{i}{N} - 0.5)\}_{i=1}^{N}$. Then, the coordinate of the point $p_i$ and the corresponding horizon-depth $d_i$ can be obtained as:
\begin{equation}
    \label{eq:coordinate}
    \begin{aligned}
    \mathbf{p}_i &= (x_i, y_i, z_i), \\
    d_i &= \sqrt{x_i^2 + y_i^2}.
    \end{aligned}
\end{equation}

The sampled points $\mathbf{P} = \{ \mathbf{p}_i\}_{i=1}^{N}$ can be converted into horizon-depth sequence $\{ d_i\}_{i=1}^{N}$. In addition, the height $h$ of each room is adopted to further supervise the model prediction on the vertical direction.


As illustrated in Figure~\ref{fig:framework}, the visual feature sequence extracted by the feature extractor $G(\cdot)$ is passed to a sequential neural networks $F(\cdot)$ to predict the horizon-depth $\hat{\mathbf{d}} = \{ \hat{d}_i \}_{i=1}^{N}$ and room height sequence $\mathbf{\hat{h}} = \{ \hat{h}_i \}_{i=1}^{N}$, respectively, through two separate branches in the output layer. A limitation of prior work is the assumption that the room height is always the same in the indoor space. Differently, we predict $\hat{h}_i$ for each sampled point $\mathbf{p}_i \in \mathbf{P}$ with ground-truth height denoted as $h_i \in \mathbf{h}$, where $\mathbf{h}=\{h_i\}_{i=1}^{N}$, which fits the design of our proposed structure variation generalization module (Introduced in Section~\ref{CSMix}). Our sequential neural networks is a SWG-Transformer module following \cite{jiang2022lgt}. The predicted horizon-depth and room height are supervised by ground-truth formulated as:
\begin{equation}
    \label{eq:loss_d}
    \begin{aligned}
        \mathcal{L}_d &= \frac{1}{N} \sum_{i=1}^{N} | d_i - \hat{d}_i|, \\
        \mathcal{L}_h &= \frac{1}{N} \sum_{i=1}^{N}|h_i - \hat{h}_i|,
    \end{aligned}
\end{equation}
where $\hat{d}_i$ and $d_i$ are the predicted and ground-truth horizon-depth at point $p_i$, respectively, and $\hat{h}_i$ and $h_i$ are the room height prediction and ground-truth, respectively.

\subsection{Appearance Variation Generalization (AVG)}
Capturing panoramas in different situations (e.g., illuminations, wall/carpet colors, cameras, and textures) would significantly change the visual appearances of the panorama images. Inspired by domain generalization problems and techniques, we propose AVG to disentangle appearance variations of panoramas during the training. This encourages the model to focus only on the important indoor structural information for the downstream task and ignore the rest. Subsequently, the model provides a better generalization capacity to novel appearances.

Specifically, the visual features $\mathbf{Z}$ extracted from $G(\cdot)$ are input to an appearance encoder $E_v(\cdot)$ which projects the visual features into a Normal distribution $\mathbf{u} \sim \mathcal{N}(0, \mathbf{I})$, then a decoder $D_v(\cdot)$ project the latent embeddings $\mathbf{u}$ into the original visual features space as $\mathbf{\hat{Z}}$ by minimizing a regularized norm-based loss:
\begin{equation}
    \begin{aligned}
        \mathcal{L}_{var} = \| \mu(\mathbf{Z}) - \mu(\mathbf{\hat{Z}})\|_2 + \| \sigma(\mathbf{Z}) - \sigma(\mathbf{\hat{Z}})\|_2.
    \end{aligned}
\end{equation}

To further enhance the generalizability of the model to novel appearance, we randomly sample latent embeddings $\mathbf{u}^{*} \sim \mathcal{N}(0, \mathbf{I})$ and project them into the visual features space as $\mathbf{Z}^{*}$, which is a novel appearance variation never observed in the training data. Then, the style information of the randomly sampled feature $\mathbf{Z}^{*}$ is transferred to the real content sample $\mathbf{Z}$ via AdaIN \cite{huang2017arbitrary} strategy as (\cite{zhou2020domain} similar to style mixup, \cite{li2021domain} Feature Stylization):
\begin{equation}
    \label{eq:adain}
    \begin{aligned}
    \mathbf{\tilde{Z}} = \sigma(\mathbf{Z}^{*})\big( \frac{\mathbf{Z} - \mu(\mathbf{Z})}{\sigma(\mathbf{Z})}\big) + \mu(\mathbf{Z}^{*}),
    \end{aligned}
\end{equation}
where the synthetic sample $\mathbf{\tilde{Z}}$ is then input to the following sequential module to predict the horizon depth and room height and optimize the model as Eq.~(\ref{eq:loss_d}) since they share the same content and spatial structure information and layout.

\begin{figure}[t]
\centering
\includegraphics[width=0.95\linewidth]{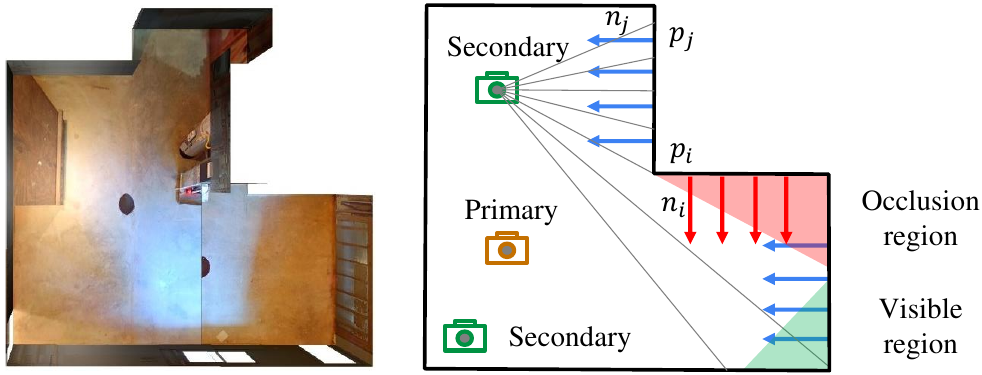}
\caption{Corners and occlusions awareness constraint.}
\label{fig:framework_gradient}
\vspace{-2mm}
\end{figure}

\subsection{Complex Structure Mix-up (CSMix)}\label{CSMix}
Another main issue that exists in prior works is overfitting to simple room shapes (e.g., Manhattan world or Atalanta world pre-assumption), and cannot handle minority and complex room shapes. Thus, in order to enhance the training data space complexity, some novel and complex samples are synthesized via cross-room, column-based mix-up strategy. Specifically, for two randomly paired samples $\{ \mathbf{Z}_i, \mathbf{Z}_j\}$, we randomly exchange columns across these two samples, which will result in two new samples with more complex structures. It is noteworthy that the synthesized samples contain more complex layout structures and also mixed appearance variations due to the cross-room mix-up. However, to avoid the random mix-up producing unreasonable noisy samples harming the optimization, a sequence of consecutive columns with width $1 <= w <= N$ is randomly selected from the two samples and mix-up, where $N$ is the width, i.e., the number of columns, of the whole visual feature maps:
\begin{equation}
    \label{eq:mixup}
    \begin{aligned}
    \mathbf{\bar{Z}}_i &= \mathbf{Z}_i[ : c_i] \oplus \mathbf{Z}_j[c_j : c_j+w] \oplus \mathbf{Z}_i[c_i + w: ], \\
    \mathbf{\bar{Z}}_j &= \mathbf{Z}_j[ : c_j] \oplus \mathbf{Z}_j[c_i : c_i+w] \oplus \mathbf{Z}_j[c_j + w: ],
    \end{aligned}
\end{equation}
where $\oplus(\cdot, \cdot)$ is concatenation operation, and $1 <= c_i <= N - w$ and $1 <= c_j <= N - w$ are randomly sampled points or columns start indices for $\mathbf{Z}_i$ and $\mathbf{Z}_j$, respectively. Similarly, the synthesized samples $\mathbf{\bar{Z}}_{i/j}$ are also input to the following layout estimation to improve the generalizability of the whole framework. It is noteworthy that since the mix-up synthesis is applied across rooms, so the room heights of the synthesized samples are also mixed up, e.g., $\mathbf{\bar{h}}_i = \mathbf{h}_i[ : c_i] \oplus \mathbf{h}_j[c_j : c_j+w] \oplus \mathbf{h}_i[c_i + w: ]$. Then the model is optimized through learning objectives as Eq.~(\ref{eq:loss_d}).

\subsection{Corners and Occlusions Awareness Constraint}

\begin{figure}[t]
\centering
\includegraphics[width=1\linewidth]{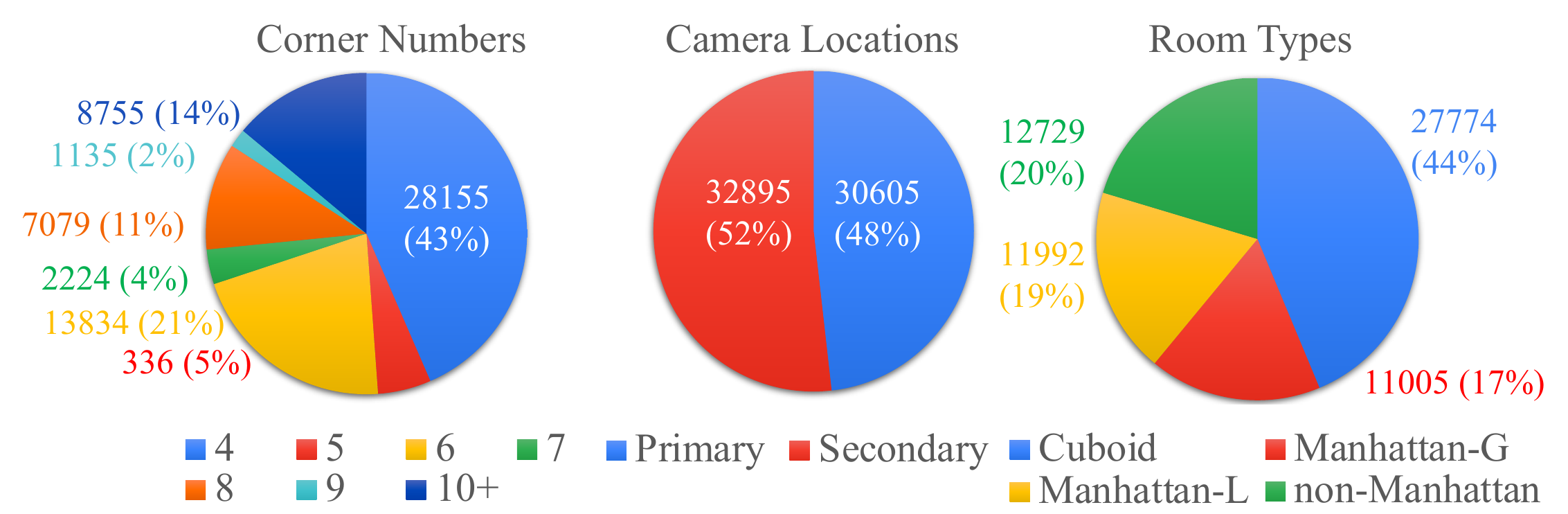}
\caption{Statistics of ZInD dataset split by different attributes. }
\label{fig:dataset_split}
\vspace{-2mm}
\end{figure}

\begin{table*}[!htb]
\caption{Estimation performance based on the number of layout corners. We can see our iBARLE achieves the highest performances in almost all metrics in all corner numbers separately. It shows the effectiveness of iBARLE for improve the performances of all cases without sacrifice some specific subsets.}
\setlength{\tabcolsep}{3.2pt}
\centering
\scalebox{0.85}{
\begin{tabular}{ccccc|cccc|cccc|cccc}
\toprule
Corner & \multicolumn{4}{c}{HorizonNet \cite{sun2019horizonnet}} & \multicolumn{4}{c}{LED$^2$-Net \cite{wang2021led}} & \multicolumn{4}{c}{LGT-Net \cite{jiang2022lgt}} & \multicolumn{4}{c}{iBARLE (Ours)} \\
\cmidrule{2-17}
Number & 2DIoU & 3DIoU & RMSE & $\delta_1$ & 2DIoU & 3DIoU & RMSE & $\delta_1$ & 2DIoU & 3DIoU & RMSE &  $\delta_1$ & 2DIoU & 3DIoU & RMSE & $\delta_1$ \\
\midrule
4   & 86.07 & 84.16 & 0.19 & \textbf{0.94} & 86.24 & 84.42 & 0.19 & 0.93 & 87.21 & 85.37 & \textbf{0.17} & \textbf{0.94} & \textbf{88.22} & \textbf{86.38} & 0.18 & \textbf{0.94} \\ 
5   & 83.76 & 80.66 & 0.28 & 0.92 & 84.49 & 82.27 & 0.25 & 0.92 & 85.76 & 83.44 & 0.22 & \textbf{0.93} & \textbf{87.83} & \textbf{85.74} & \textbf{0.20} & \textbf{0.93} \\ 
6   & 83.73 & 81.78 & 0.22 & 0.93 & 83.29 & 81.32 & 0.22 & 0.93 & 83.50 & 81.66 & 0.21 & 0.93 & \textbf{85.50} & \textbf{83.57} & \textbf{0.19} & \textbf{0.94} \\ 
7   & 76.02 & 73.54 & 0.30 & 0.90 & 77.45 & 73.86 & 0.29 & 0.89 & \textbf{79.68} & \textbf{77.22} & 0.28 & 0.91 & 79.62 & 76.92 & \textbf{0.25} & \textbf{0.92} \\ 
8   & 80.38 & 78.18 & 0.22 & 0.93 & 79.97 & 77.72 & 0.22 & 0.93 & 80.13 & 77.98 & 0.23 & 0.92 & \textbf{80.69} & \textbf{78.55} & \textbf{0.20} & \textbf{0.94} \\ 
9   & \textbf{81.92} & \textbf{79.59} & 0.25 & 0.92 & 81.65 & 79.17 & 0.25 & 0.91 & 80.39 & 78.17 & 0.26 & 0.92 & 81.14 & 78.75 & \textbf{0.23} & \textbf{0.93}\\ 
$10+$   & 75.42 & 72.61 & 0.30 & 0.91 & 74.75 & 71.91 & 0.29 & 0.91 & 75.21 & 72.26 & 0.29 & 0.90 & \textbf{76.16} & \textbf{73.39} & \textbf{0.25} & \textbf{0.92}\\ 
\midrule
Average   & 81.04 & 78.65 & 0.25 & 0.92 & 81.12 & 78.67 & 0.24 & 0.92 & 81.70 & 79.44 & 0.24 & 0.92 & \textbf{82.74} & \textbf{80.47} & \textbf{0.21} & \textbf{0.93}\\ 
\bottomrule
\end{tabular}
}
\label{table:result_corner_number}
\end{table*}

\begin{table*}[!thb]
\caption{Results comparison on ZInD dataset split by type of layout. Results on each group with specific type of layout are reported separately, and the group-wise average results are also compared.
}

\setlength{\tabcolsep}{2.5pt}
\centering
\scalebox{0.85}{
\begin{tabular}{ccccc|cccc|cccc|cccc}
\toprule
Room & \multicolumn{4}{c}{HorizonNet \cite{sun2019horizonnet}} & \multicolumn{4}{c}{LED$^2$-Net \cite{wang2021led}} & \multicolumn{4}{c}{LGT-Net \cite{jiang2022lgt}} & \multicolumn{4}{c}{iBARLE (Ours)} \\
\cmidrule{2-17}
Type & 2DIoU & 3DIoU & RMSE & $\delta_1$ & 2DIoU & 3DIoU & RMSE & $\delta_1$ & 2DIoU & 3DIoU & RMSE &  $\delta_1$ & 2DIoU & 3DIoU & RMSE & $\delta_1$ \\
\midrule
Cuboid & 86.47 & 84.54 & 0.19 & \textbf{0.94} & 86.63 & 84.78 & 0.19 & \textbf{0.94} & 87.54 & 85.69 & \textbf{0.17} & \textbf{0.94} & \textbf{88.62} & \textbf{86.76} & 0.18 & \textbf{0.94}\\ 
Manhattan-l   & 83.50 & 81.61 & 0.21 & 0.93 & 83.03 & 81.15 & 0.22 & 0.93 & 83.29 & 81.43 & 0.21 & 0.93 & \textbf{85.13} & \textbf{83.21} & \textbf{0.19} & \textbf{0.94} \\ 
Manhattan-g    & 78.08 & 76.00 & 0.25 & 0.92 & 77.66 & 75.45 & 0.24 & 0.92 & 78.19 & 75.90 & 0.25 & 0.92 & \textbf{78.90} & \textbf{76.89} & \textbf{0.21} & \textbf{0.93} \\ 
non-Manhattan   & 79.99 & \textbf{76.97} & 0.28 & 0.91 & 80.18 & 77.23 & 0.27 & 0.91 & 80.83 & 78.33 & 0.25 & 0.92 & \textbf{82.08} & 79.36 & \textbf{0.23} & \textbf{0.93} \\ 
\midrule
 Average   & 82.01 & 79.78 & 0.23 & 0.93 & 81.87 & 79.66 & 0.23 & 0.92 & 82.46 & 80.34 & 0.22 & 0.93 & \textbf{83.68} & \textbf{81.55} & \textbf{0.20} & \textbf{0.94} \\ 
\bottomrule
\end{tabular}
}
\label{table:result_room_type}
\end{table*}

As shown in Figure~\ref{fig:framework_gradient}, each wall is a plane but the positions on the same wall could have different horizon depths. Thus, to supervise the planar of the walls, the normals at different positions of the same wall are constrained consistently \cite{jiang2022lgt}. Since the wall and floors are preassumed perpendicular to the floor and the normals. Thus we first convert the ground-truth and predicted horizon depth at each position back into the 3D point $\mathbf{p}_i = (d_i \sin(\theta_i), h^f, d_i \cos(\theta_i))$ and $\mathbf{\hat{p}}_i = (\hat{d}_i \sin(\theta_i), h^f, \hat{d}_i \cos(\theta_i))$, where $h^f$ is the height from the camera center to the floor and $\theta_i$ is the angle of the logitude. It it noteworthy that since the expected normal vectors are parallel to the floor, thus the height of both $\mathbf{p}_i$ and predicted $\mathbf{\hat{p}}_i$ are set as $h^f$, which will not influence the computation of normal vectors. Then ground-truth and predicted normal vectors at the same position are computed as:
\begin{equation}
    \label{eq:normal}
    \begin{aligned}
    &\mathbf{n}_i = \mathbf{M}_r \Big(\frac{\mathbf{p}_{i+1} - \mathbf{p}_i}{\|\mathbf{p}_{i+1} - \mathbf{p}_i\|}\Big)^{\top},\\
    &\mathbf{\hat{n}}_i = \mathbf{M}_r \Big(\frac{\mathbf{\hat{p}}_{i+1} - \mathbf{\hat{p}}_i}{\|\mathbf{\hat{p}}_{i+1} - \mathbf{\hat{p}}_i\|}\Big)^{\top}, 
    \end{aligned}
\end{equation}
where $\mathbf{M}_r$ is the rotation matrix of $\frac{\mathbf{p}_i}{2}$, and $\mathbf{n}_i$ and $\mathbf{hat{n}}_i$ are the ground-truth and predicted normal vectors at the same position, respectively. The learning objective is defined as maximizing inner product between the predicted and ground-truth normals as:
\begin{equation}
    \label{eq:loss_normal}
    \begin{aligned}
    \mathcal{L}_{n} = \frac{1}{N} \sum_{i=1}^{N}|-\mathbf{n}_i \cdot \mathbf{\hat{n}}_i|.
    \end{aligned}
\end{equation}

Moreover, the normals change near the corners, thus the gradient of the normal angles are obtained to supervise the turning of corners \cite{jiang2022lgt}. However, for complex indoor space beyond simple cuboid or Manhattan world, occlusions appear and the normals near the boundaries remain consistent but the depth change sharply. In order to capture the changing of visible corners and invisible boundaries near occlusions, the gradient constraints to both the normal vectors and depth prediction are applied. Specifically, the gradient of the normals and depth are calculated as:
\begin{equation}
    \label{eq:loss_normal}
    \begin{aligned}
    \mathbf{g}_i^n &= \arccos (\mathbf{n}_i \cdot \mathbf{n}_{i+1}), \,\,\,
    g_i^d = d_{i+1} - d_{i}, \\
    \hat{\mathbf{g}}_i^n &= \arccos (\mathbf{\hat{n}}_i \cdot \mathbf{\hat{n}}_{i+1}), \,\,\,
    \hat{g}_i^d = \hat{d}_{i+1} - \hat{d}_{i},
    \end{aligned}
\end{equation}
where $\mathbf{n}_i / \mathbf{\hat{n}}_i$ and $d_i / \hat{d}_i$ are the ground truth and predicted normal/depth, respectively. Then, the gradient-based normal and depth prediction constraint is defined as:
\begin{equation}
    \label{eq:loss_normal}
    \begin{aligned}
    \mathcal{L}_g = \frac{1}{N}\sum_{i=1}^{N} (|\mathbf{g}_i^n - \hat{\mathbf{g}}_i^n| + |g_i^d - \hat{g}_i^d|).
    \end{aligned}
\end{equation}

To this end, the aggregated learning objective of layout estimation with the panorama images as input is obtained via integrating the aforementioned losses as $\mathcal{L}(\mathbf{Z}) = \mathcal{L}_d + \mathcal{L}_h + \mathcal{L}_n + \mathcal{L}_g$, where $\mathbf{Z}$ denotes the visual features extracted by $G(\cdot)$ with panorama images as input.

\begin{table*}[!htb]
\caption{Estimation performance based on different panorama capture locations. Our iBARLE achieves the highest performances for all primary/secondary categories and the overall performance.}
\setlength{\tabcolsep}{3pt}
\centering
\scalebox{0.85}{
\begin{tabular}{ccccc|cccc|cccc|cccc}
\toprule
Camera & \multicolumn{4}{c}{HorizonNet \cite{sun2019horizonnet}} & \multicolumn{4}{c}{LED$^2$-Net \cite{wang2021led}} & \multicolumn{4}{c}{LGT-Net \cite{jiang2022lgt}} & \multicolumn{4}{c}{iBARLE (Ours)} \\
\cmidrule{2-17}
Pose & 2DIoU & 3DIoU & RMSE & $\delta_1$ & 2DIoU & 3DIoU & RMSE & $\delta_1$ & 2DIoU & 3DIoU & RMSE &  $\delta_1$ & 2DIoU & 3DIoU & RMSE & $\delta_1$ \\
\midrule

Primary   & 85.58 & 83.71 & 0.21 & 0.94 & 85.93 & 84.00 & 0.20 & \textbf{0.93} & 86.23 & 84.41 & \textbf{0.19} & 0.94 & \textbf{87.72} & \textbf{85.85} & \textbf{0.19} & 0.94\\ 
Secondary   & 81.16 & 78.81 & 0.23 & 0.93 & 80.70 & 78.46 & 0.23 & \textbf{0.92} & 81.57 & 79.33 & 0.22 & 0.93 & \textbf{82.63} & \textbf{80.44} & \textbf{0.20} & \textbf0.93 \\ 
\midrule
Average   & 83.37 & 81.26 & 0.22 & 0.93 & 83.32 & 81.23 & 0.22 & 0.93 & 83.90 & 81.87 & 0.21 & 0.93 & \textbf{85.18} & \textbf{83.15} & \textbf{0.19} & \textbf{0.94} \\ 
\bottomrule
\end{tabular}
}
\label{table:result_pose}
\end{table*}

\subsection{Overall Training Objective}
Combining the horizon depth prediction, room height prediction, normals prediction, and gradient-based prediction constraint loss for all real train samples and synthesized data, produced by appearance variation domain generalization and cross-room structure mix-up, the overall learning objective of our proposed model is shown below: 
\begin{equation}
\begin{aligned}
&\underset{G, F}{\min}\  \mathcal{L}(\mathbf{Z}) + \alpha \mathcal{L}(\mathbf{\tilde{Z}}) + \beta \mathcal{L}(\mathbf{\bar{Z}}), \\
&\underset{E_v, D_v}{\min} \mathcal{L}_{var},
\end{aligned}
\end{equation}
where $\alpha$ and $\beta$ are hyper-parameters to balance the contribution of the real data, and the synthesized samples via AVG and CSMix, respectively. Moreover, the networks $E_v(\cdot)$ and $D_v(\cdot)$ in AVG are trained separately and fixed during the training of other networks.

\section{Experiment}\label{sec:experiment}

\subsection{Experimental settings}
\noindent \textbf{Datasets:} Our experiments are based on two variants: (1) Zillow Indoor Dataset (ZInD)~\cite{cruz2021zillow} is the largest indoor dataset consisting of 67,448 panorama images with room layout annotations including various and complex indoor spaces from general Manhattan, non-Manhattan, and non-flat ceilings layouts. We follow the official training/validation/test splits and adopt the ``raw'' layout annotations as ground truth. (2) ZInD-Simple~\cite{cruz2021zillow} is a subset of ZInD dataset with only simple indoor cuboid layouts without any contiguous occluded corners exist. We evaluate our proposed model on ZInD-simple to compare it with the prior state-of-the-art method.

\noindent \textbf{Data Splits:} To evaluate iBARLE on different imbalanced subsets, we split the whole test layouts into several groups with different standards as mentioned earlier. The statistics of the splits based on corner numbers, room types, and camera poses are shown in Figure~\ref{fig:dataset_split}. We evaluate the performance of our iBARLE model for each group.

\begin{table}[!t]
\caption{Overall experimental results comparison on ZInD dataset}
\setlength{\tabcolsep}{3pt}
\centering
\scalebox{0.85}{
\begin{tabular}{ccccc}
\toprule
Corner Number & 2DIoU($\%$) \color{green}{$\uparrow$} & 3DIoU($\%$) \color{green}{$\uparrow$} & RMSE \color{red}{$\downarrow$} & $\delta_1$ \color{green}{$\uparrow$} \\
\midrule
HorizonNet \cite{sun2019horizonnet} & 83.25 & 81.13 & 0.2219 & 0.9303 \\
LED$^2$-Net \cite{wang2021led} & 83.18 & 81.08 & 0.2172 & 0.9273 \\
LGT-Net \cite{jiang2022lgt}& 83.81 & 81.77 & 0.2074 & 0.9309 \\
\midrule
iBARLE (Ours) & \textbf{85.04} & \textbf{83.00} & \textbf{0.1949} & \textbf{0.9375} \\
\bottomrule
\end{tabular}
}
\label{table:result_zind_overall}
\end{table}

\begin{table}[!t]
\caption{Overall experimental results comparison on ZInD-simple}

\setlength{\tabcolsep}{3pt}
\centering
\scalebox{0.85}{
\begin{tabular}{ccccc}
\toprule
Corner Number & 2DIoU($\%$) \color{green}{$\uparrow$} & 3DIoU($\%$) \color{green}{$\uparrow$} & RMSE \color{red}{$\downarrow$} & $\delta_1$ \color{green}{$\uparrow$} \\
\midrule
HorizonNet \cite{sun2019horizonnet} & 90.44 & 88.59 & 0.123 & 0.957\\
LED$^2$-Net \cite{wang2021led} & 90.36 & 88.49 & 0.124 & 0.955\\
LGT-Net [ViT] \cite{jiang2022lgt} & 88.93 & 86.19 &0.146 & 0.950 \\
LGT-Net \cite{jiang2022lgt} & 91.77 & 89.95 & 0.111 & 0.960\\
\midrule
iBARLE (Ours) & \textbf{92.22} & \textbf{90.42} & \textbf{0.107} & \textbf{0.962} \\
\bottomrule
\end{tabular}
}
\label{table:result_zind_simple_overall}
\vspace{-2mm}
\end{table}

\begin{figure*}[t!]
\centering
\includegraphics[width=0.95\linewidth]{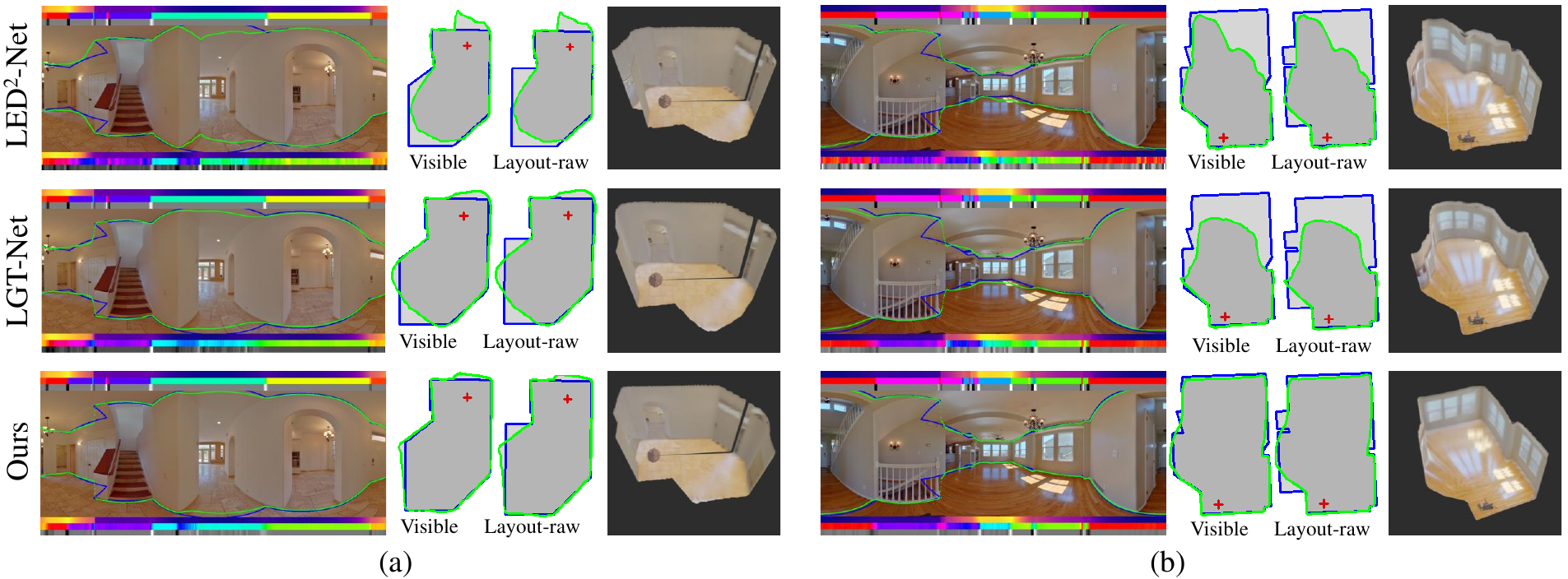}
\vspace{-1mm}
\caption{Case study: Layout estimation on samples where occlusions occur and some spaces are invisible to the panorama camera. It leads to more complex, unique, and imbalanced layout shapes. ``Visible'' denotes the layout of visible regions. ``layout-raw'' is the real-world room layout which includes the occlusion spaces. Our iBARLE model performs better for imbalanced/minority cases.}
\label{fig:qualitative_occlusion}
\vspace{-1mm}
\end{figure*}

\noindent \textbf{Evaluation Metrics:} We use four widely used metrics for evaluation: (1) 2D IoU: Intersection over the Union of 2D room layouts. (2) 3D IoU: Intersection over Union of 3D room layouts. (3) RMSE: root mean squared error of the depth prediction with the camera height as $1.6$ meters. (4) $\delta_1$: percentage of pixels where the ratio between the predicted depth and ground-truth depth is within a threshold of $1.25$~\cite{jiang2022lgt, zou2021manhattan}. For 2DIoU, 3DIoU, and $\delta_1$ metrics, higher is better which is denoted by \textcolor{green}{$\uparrow$}. On the contrary, RMSE metric is a negatively-oriented score, thus lower is better denoted by \textcolor{red}{$\downarrow$}. In addition to the overall performance based on these evaluation metrics on the test data, we also report the average results across different sub-groups split by specific standards to evaluate how balanced our model on the data space across simple to complex indoor spaces.

\begin{figure*}[t!]
\centering
\includegraphics[width=0.95\linewidth]{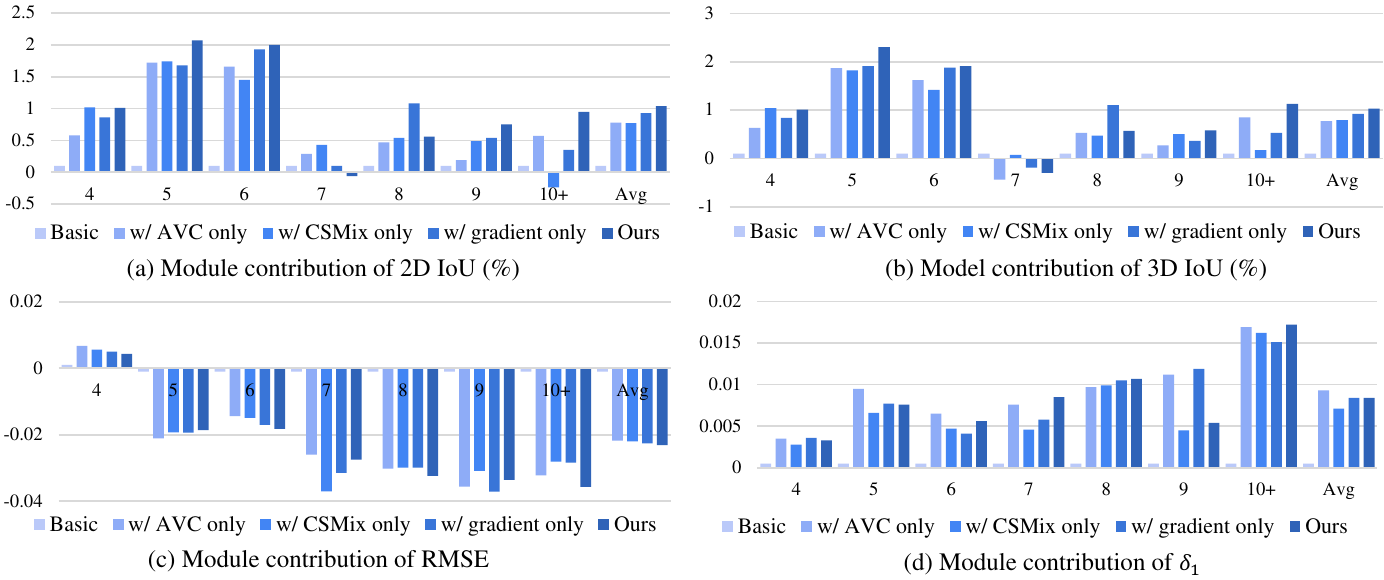}
\vspace{-1mm}
\caption{Ablation study of the module contributions to the iBARLE framework, where each module is separately added to the basic model. The 4 plots correspond to the 4 evaluation metrics. We can observe that each module is effective at improving the prediction performance. And in most of the cases, the complete model achieves the highest performance.}
\label{fig:result_zind_ablation}
\end{figure*}

\noindent \textbf{Comparison with Baselines:}
We compare our layout estimation results with those of state-of-the-art baselines, namely, HorizonNet (CVPR'19)~\cite{sun2019horizonnet}, LED$^2$-Net (CVPR'21)~\cite{wang2021led}, and LGT-Net (CVPR'22)~\cite{jiang2022lgt}.


\begin{figure*}[t!]
\centering
\includegraphics[width=0.95\linewidth]{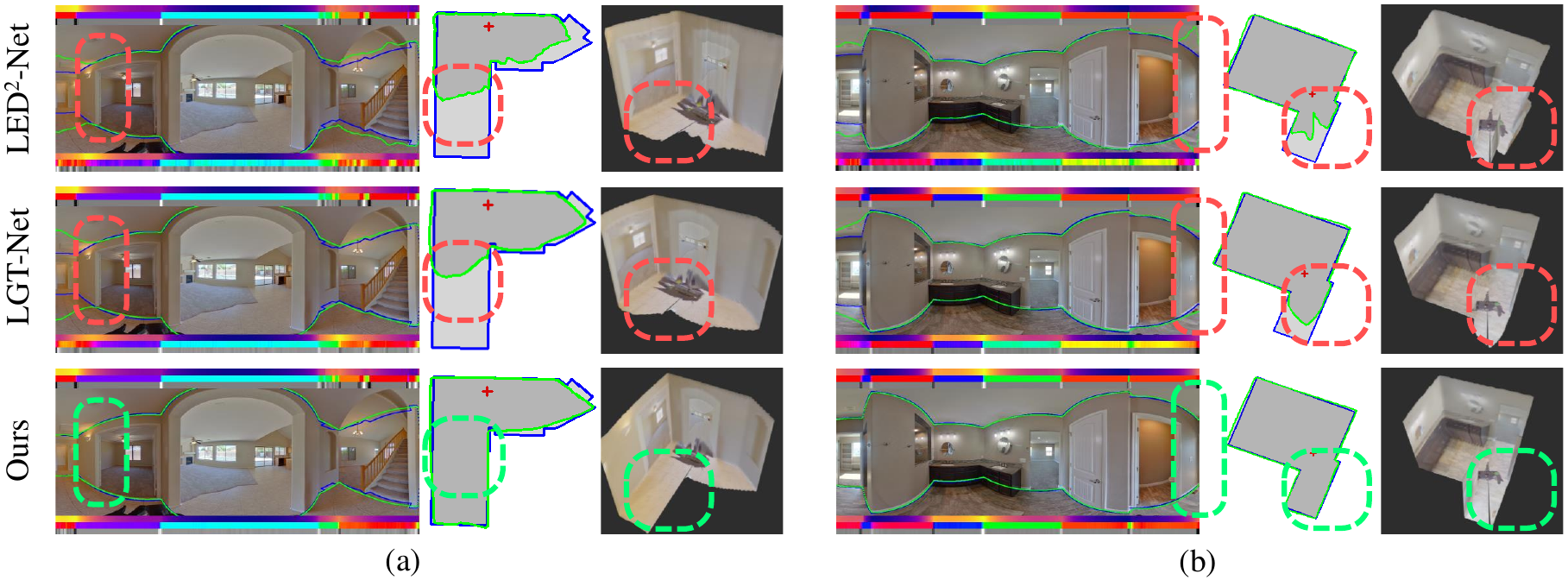}
\vspace{-1mm}
\caption{Case study: Layout estimation results beyond Manhanttan world. Our iBARLE approach predicts non-Manhanttan regions (green circle) in higher accuracy than other benchmarks. It denotes the robustness of iBARLE for handling minority cases.}
\label{fig:qualitative_pose}
\vspace{-1mm}
\end{figure*}

\subsection{Implementation Details}
In our experiments, the feature extractor uses the architecture proposed in HorizonNet \cite{sun2019horizonnet} based on ResNet-50 \cite{he2016deep}. The architecture takes a panorama with the dimension of $512 \times 1024 \times 3$ (height, width, channel) as input and gets $2D$ feature maps of 4 different scales by ResNet-50. Then, it compresses the height and up samples width $N$ of each feature map to get $1D$ feature sequences with the same dimension $\mathbb{R}^{N \times \frac{D}{4}}$ and connects them, finally outputs a feature sequence $\mathbb{R}^{N \times D}$, where $D=1024$ and $N=256$ in our implementation. For the sequential depth prediction module, we use the SWG-Transformer proposed in \cite{jiang2022lgt} which is based on Transformer \cite{vaswani2017attention}. The whole framework is implemented with PyTorch and optimized by Adam optimizer with a learning rate set as $1e^{-4}$. For hyper-parameters, we empirically fix $\alpha=0.1, \beta=0.01$. The hyper-parameters of the layout estimation objectives in $\mathcal{L}(\mathbf{Z})$ follow the same setting as \cite{jiang2022lgt} for a fair comparison.

\subsection{Results Comparison}
\noindent \textbf{Layout Estimation across Imbalanced Data.} The results on groups with a different number of corners and different room shapes are shown in Table~\ref{table:result_corner_number} and Table~\ref{table:result_room_type}, respectively. We notice the performance degradation from simple shape rooms to complex spaces, which demonstrates the motivation of exploring the imbalance issues of layout estimation across arbitrary structures. From the results, we observe that our proposed method outperforms state-of-the-art baselines in most cases with various metrics. More specifically, for the group-wise average 2D IoU and 3D IoU calculated across different numbers of corners, our model outperforms the second-best baseline by $1.04\%$ and $1.03\%$, respectively. From the results in Table~\ref{table:result_room_type}, our iBARLE model improves the 2D IoU on Manhattan-l and non-Manhattan groups both over $1.5\%$ compared to LGT-Net.

Moreover, we split the ZInD dataset into subsets based on the location where the camera taking the panoramas. The results are shown in Table~\ref{table:result_pose}. Based on the definition of the camera pose of ZInD, panoramas that are taken with ``Primary'' pose capture more content and are easier to capture the whole view of the room. On the contrary, images taken with the ``Secondary'' pose usually contain less information and are harder to estimate the layout since occlusions are more possible to occur and the camera could be too close/far from some walls. From the results, we observe our model outperforms all compared baselines on different groups and beats the second-best baseline more than $1.0\%$ for average 2D IoU and 3D IoU across the types of camera pose. 

\noindent \textbf{Overall Layout Estimation.}
To compare the overall layout estimation performance with prior layout estimation baselines, we report the overall results on ZInD dataset in Table~\ref{table:result_zind_overall}. From the results, we observe that our proposed framework achieves a new state-of-the-art layout estimation performance for all metrics. iBARLE improves the overall 2D IoU by $1.23\%$ over LGT-Net. Furthermore, most prior layout estimation solutions are designed for simple room shapes, e.g., cuboids. Thus, we apply our proposed model to ZInD-simple dataset, which is a subset of ZInD consisting of only simple cuboid layouts without any occluded corners exist. From the results in Table~\ref{table:result_zind_simple_overall}, we observe that our model can beat the state-of-the-art baselines on simple layout subsets. Although the performance improvement on the simpler layout shape ZInD-simple subset is less significant than on the more complex and diverse ZInD dataset, the results in Table~\ref{table:result_zind_simple_overall} demonstrate the effectiveness of the designed modules in iBARLE.

\section{Discussion}

\paragraph{Ablation Study.}
We further conduct an ablation study to evaluate the contribution of each module in iBARLE. Specifically, we compare the performance changes, increases of 2D IoU/3D IoU/$\delta_1$ and decreases of RMSE, when each of these modules is stacked to the basic model. The results changes of the four metrics are illustrated in Figure~\ref{fig:result_zind_ablation}. From the results, we observe that all three modules are able to effectively enhance the estimation performance on almost all subgroups. It demonstrates the contribution and effectiveness of the designed three modules for layout estimation, especially for the imbalanced scenario. Moreover, our complete iBARLE model with all modules aggregated can achieve the highest performance which denotes the smoothness of the whole model structure.  

\paragraph{Qualitative Analysis.}
To intuitively check the effectiveness of our iBARLE model, we visualize some results on the ZInD dataset in Figure~\ref{fig:qualitative_occlusion} and Figure~\ref{fig:qualitative_pose}. The boundaries of the room layout are displayed on the panorama and the floor plan. The blue lines are the ground truth, the green lines are predictions, and the red cross is the position of the camera. In addition, the predicted horizon depth, normal, and gradient of the normal are visualized as heat maps under the panorama image, and the ground truth is shown on top of the image. Moreover, 3D visualization of the selected samples is reconstructed based on the predicted layout, and the red dash lines highlight the errors made by the compared baselines. From the results in Figure~\ref{fig:qualitative_occlusion}, we observe that our model can manage the layout estimation of complex rooms beyond the Manhattan world assumption. 
Moreover, we also observe the difference between the ``raw layout'' and the layout ``visible'' to the camera. Our model can predict accurately with occlusion corners.
Besides, results shown in Figure~\ref{fig:qualitative_pose} are with challenging camera poses, e.g., close to the wall or corner. From the results, we can observe the proposed model is robust to predict layout with complex space panoramas taken on arbitrary positions in the room. . 

\paragraph{Non-Manhattan Results.}
From Table~\ref{table:result_room_type}, our layout estimation performance on the ``Non-Manhattan'' group is not as significant as other groups. Our hypothesis is that the ``Non-Manhattan'' group contains samples that have far more complex and diverse structures. In addition, the number and positions of the selected columns in the CSMix module are hyper-parameters influencing the complexity of the synthesized samples. These issues require more in-depth investigation on generalizability and robustness.

\section{Conclusion}\label{sec:conclusion}
We propose a new technique, iBARLE, to address problems associated with data imbalance and appearance variation for single-image room layout estimation. The key components are the Appearance Variation Generalization (AVG) and Complex Structure Mix-up (CSMix) modules, which help generalize over a wide range of complex and diverse room shapes as well as scene appearance variations. We also use a gradient-based layout estimation constraint to account for occlusions in complex layouts.
Experimental results show that iBARLE improves room layout estimation across different kinds of imbalanced distributions of majority and minority groups in the dataset. iBARLE also improves overall layout estimation performance.


\balance
{\small
\bibliographystyle{ieee_fullname}
\bibliography{egbib}
}

\end{document}